\documentclass[11pt]{article}
\usepackage[margin=1in]{geometry}
\usepackage[numbers,sort&compress]{natbib}
\usepackage{amsmath,amssymb}
\usepackage{graphicx}
\usepackage{booktabs}
\usepackage{multirow}
\usepackage{arydshln}
\usepackage{placeins}
\usepackage{hyperref}
\usepackage{url}
\usepackage{listings}
\usepackage[most]{tcolorbox}
\usepackage{ragged2e}

\title{SLICE-Mem: Salience-Gated Lightweight Information Curation for Efficient LLM Agent Memory}\author{%
  Chen Xiaobin$^{1}$, Jin Dian$^{1}$, Shu Lin$^{2}$,\\
  Sun Defeng$^{3}$, and Tao Xiaoming$^{1}$\\[0.5em]
  \small
  $^{1}$School of Fashion and Textiles, The Hong Kong Polytechnic University\\
  $^{2}$School of Future Technology, South China University of Technology\\
  $^{3}$Department of Applied Mathematics, The Hong Kong Polytechnic University
}
\date{}

\begin{document}
\maketitle

\begin{abstract}
Long-term memory enables large language model (LLM) agents to maintain personalized and temporally coherent interactions, but repeated LLM inference for memory admission, extraction, and organization introduces substantial cost and latency. We present SLICE-Mem, a lightweight memory framework that combines salience-guided admission and predicate-centered information extraction. A compact classifier identifies user turns worth retaining, and a semantic-dependency-based extractor converts admitted turns into concise memory entries. This online pipeline requires no generative LLM calls, while cross-entry reconciliation is deferred to an offline consolidation stage. We evaluate SLICE-Mem on the weekly subset of Memora and LongMemEval-S using 8B and 30B Qwen backbones for memory construction and consolidation on both benchmarks. Compared with the generative-memory baseline MemOS under the Qwen3-30B configuration, SLICE-Mem’s online pipeline achieves 46.60 versus 46.68 FAMA (Forgetting-Aware Memory Accuracy) on Memora Weekly and 56.8\% versus 59.3\% accuracy on LongMemEval-S, while reducing total LLM token consumption by 92.4\% and 99.5\%, respectively. These results demonstrate that SLICE-Mem can preserve useful evidence for long-term memory with substantially less reliance on LLM-based memory construction.
\end{abstract}

\section{Introduction}

Long-term memory is essential for large language model (LLM) agents that interact with users across sessions. Such agents must retain preferences, past events, commitments, and evolving constraints to provide coherent and personalized assistance. Yet the underlying LLM is effectively stateless across independent sessions, and a finite context window cannot serve as durable memory. Agent systems therefore externalize interaction histories into persistent stores \citep{packer2023memgpt,park2023generative,zhong2024memorybank}. As these histories grow, memory construction becomes a bottleneck: many turns are transient or redundant, indiscriminate storage introduces retrieval noise, and repeatedly invoking LLMs for filtering, extraction, and organization incurs substantial latency and token cost. Efficient long-term memory must therefore determine, before expensive processing, which information merits retention and how it can be represented with minimal online computation.

Existing memory systems primarily optimize downstream retrieval or answer accuracy. Structured and hierarchical approaches improve organization \citep{kang2025memoryos,xu2025amem}, but often call a general-purpose LLM throughout the memory lifecycle to select interactions, extract facts, detect conflicts, and consolidate records. Retrieval-augmented generation supplies external evidence at answer time \citep{lewis2020rag}, but long-context use remains challenging \citep{bai2024longbench,liu2024lost}; neither directly controls admission to persistent memory. Hierarchical and graph-based stores likewise rely on LLM-generated summaries, graphs, or atomic units \citep{gutierrez2024hipporag,kang2025memoryos,sarthi2024raptor,xu2025amem}. Recent lightweight designs reduce repeated summarization or defer maintenance to an offline phase \citep{fang2026lightmem,hu2025hiagent}, but early admission and non-generative extraction remain underexplored as a unified construction strategy. This gap affects both efficiency and reliability: irrelevant or stale entries enlarge the retrieval space and may steer later agent behavior \citep{xiong2026memorymanagement}.

We introduce \textbf{SLICE-Mem}, a \textbf{S}alience-gated \textbf{L}ightweight \textbf{I}nformation \textbf{C}uration framework for \textbf{E}fficient agent memory. Inspired by selective encoding and sleep-dependent consolidation \citep{atkinson1968human,diekelmann2010memory,rasch2013sleep}, SLICE-Mem separates lightweight online formation from deferred cross-entry reconciliation. A compact Salience Gate learns to identify user turns worth retaining from consensus labels produced by three heterogeneous LLM teachers and released memory-admission annotations. By considering each target turn together with its preceding dialogue context, the Gate filters low-value content before parsing and embedding. For admitted turns, a deterministic extractor uses semantic dependencies to identify predicates and their associated arguments and modifiers, then assembles the corresponding text spans into concise memory entries. This process preserves surface evidence and turn-level provenance without generative LLM calls during online admission or extraction. Offline consolidation subsequently addresses redundancy and conflicting information through relation classification and entry-status updates while preserving the original memory text. We evaluate SLICE-Mem on Memora Weekly and LongMemEval-S using 8B and 30B Qwen backbones for memory construction and consolidation, with answer generation and judging held fixed. In the 30B comparison, the online pipeline achieves 46.60 versus 46.68 FAMA on Memora Weekly and 56.8\% versus 59.3\% accuracy on LongMemEval-S relative to the generative-memory baseline MemOS~\citep{li2025memos}. It reduces total LLM token consumption by 92.4\% and 99.5\% and reported memory-construction time by 84.8\% and 61.0\%, respectively. Separate evaluation of the online and consolidated states reveals task- and backbone-dependent effects of consolidation. Together, these results demonstrate a favorable effectiveness–cost trade-off for selective, non-generative memory formation while highlighting the need for evidence-preserving consolidation.

Our main contributions are threefold:
\begin{itemize}
    \item We develop a compact Salience Gate that learns context-aware memory admission from consensus-based LLM annotations and released admission labels, filtering user turns before extraction and embedding without generative LLM inference at deployment.
    
    \item We design a deterministic, semantic-dependency-guided extractor that converts admitted turns into concise memory entries while preserving surface evidence and turn-level provenance. Together with the Gate, it enables online memory formation without generative LLM calls.
    
    \item We evaluate SLICE-Mem on two long-term memory benchmarks with 8B and 30B Qwen backbones, demonstrating competitive answer quality at substantially lower token and time cost, and identifying the task-dependent benefits and limitations of offline consolidation.
\end{itemize}

\begin{figure}[!t]
  \centering
  \includegraphics[width=0.9\textwidth]{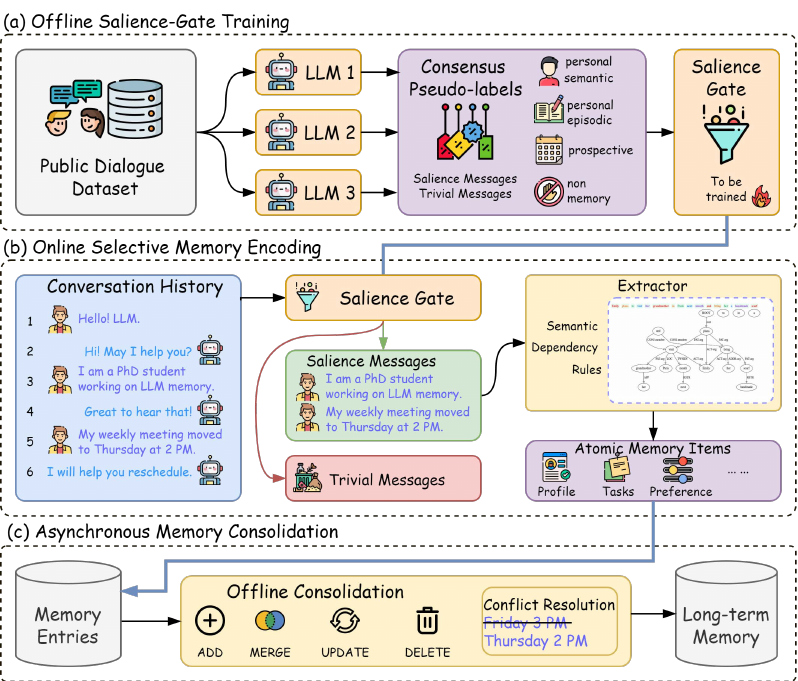}
  \caption{Overview of SLICE-Mem architecture. (a) Consensus-based salience gate training. (b) Dependency-guided extraction and turn-level memory storage. (c) Offline reconciliation of memory entries.}
  \label{fig:framework}
\end{figure}

\section{Related Work}

\paragraph{Memory architectures.}
Persistent-memory agents differ primarily in how they represent and organize past interactions. Generative Agents retrieve chronological memories using relevance, recency, and LLM-estimated importance, and periodically synthesize reflections \citep{park2023generative}. MemoryBank maintains evolving user profiles with time-dependent forgetting \citep{zhong2024memorybank}, whereas MemGPT treats the context window as working memory and moves information across virtual tiers \citep{packer2023memgpt}. Later systems introduce more structured organization: HiAgent compresses completed subgoals \citep{hu2025hiagent}, A-MEM maintains linked atomic notes \citep{xu2025amem}, and MemoryOS applies tier-specific update policies \citep{kang2025memoryos}. These designs improve retrieval and multi-session reasoning, but commonly rely on repeated LLM inference for selection, reflection, summarization, or note construction.

\paragraph{Efficient and adaptive management.}
Efficiency-oriented methods reduce either the context presented to an agent or the cost of maintaining its store. LightMem moves expensive consolidation outside the online path and avoids repeatedly processing the full history \citep{fang2026lightmem}. Learned managers instead optimize memory operations: AgeMem trains a unified policy over storing, retrieving, updating, summarizing, and discarding \citep{yu2026agemem}, while Memory-R1 learns \textsc{Add}, \textsc{Update}, \textsc{Delete}, and \textsc{Noop} decisions jointly with memory-aware answering \citep{yan2026memoryr1}. Although adaptive, these approaches require policy training and often retain generative models in the management loop. Moreover, efficient retrieval cannot repair poor entries after ingestion, since agents may follow retrieved experiences even when they are erroneous \citep{xiong2026memorymanagement}.

\paragraph{Salience-aware curation and lightweight extraction.}
Selective memory is usually implemented through scalar importance scores, heuristic retention rules, or prompt-based LLM judgments. Generative Agents combine importance with recency and relevance at retrieval time \citep{park2023generative}, while MemoryBank couples importance with forgetting \citep{zhong2024memorybank}. Such mechanisms may conflate different reasons for remembering, including future utility, persistent personal relevance, and social significance. Structured systems also commonly ask an LLM to convert interactions into summaries, profiles, or atomic notes before pruning or consolidation \citep{kang2025memoryos,xu2025amem}. In contrast, SLICE-Mem makes an explicit admission decision before extraction and uses deterministic linguistic processing for admitted turns, removing LLM calls from online memory construction. Its contribution is therefore complementary to hierarchical storage and learned lifecycle management. Related structured, temporal, linguistic, and admission-control designs include EnSIMem, TSM, Semantic Anchoring, and A-MAC \citep{meng2026ensimem,su2026tsm,chatterjee2025anchoring,zhang2026amac}; Appendix~\ref{app:related_scope} contrasts their mechanisms with ours.

\section{Method}
\label{sec:method}

As illustrated in Figure~\ref{fig:framework}, SLICE-Mem separates online memory formation from offline consolidation. Given a dialogue $U=(u_1,\ldots,u_T)$, with speaker, text, and timestamp for each turn, the online pipeline admits informative user turns, extracts predicate-centered statements, and inserts their embeddings with timestamp and provenance. Admission and extraction use a compact classifier and deterministic linguistic rules without generative LLM calls. Consolidation subsequently reconciles relations among stored entries without rewriting their text.

\subsection{Salience-Gated Memory Admission}
\label{sec:gate}
The Gate predicts whether retaining a user turn may support future continuity, personalization, factual consistency, or plan completion. For a target user turn $u_t$, let $H_t^{(k)}$ contain up to $k$ immediately preceding dialogue messages. We construct $z_t=\operatorname{Pack}(H_t^{(k)},u_t)$, using explicit markers to distinguish the context from the target. The Gate encodes this input and predicts admission through a single-logit classification head:
\begin{equation}
 p_t=\sigma\!\left(\mathbf{w}^{\top}\operatorname{Enc}_{\theta}(z_t)+b\right),
 \qquad g(u_t)=\mathbf{1}[p_t\geq\tau_g].
 \label{eq:gate_admission}
\end{equation}
Here, $\operatorname{Enc}_{\theta}(z_t)\in\mathbb{R}^{d}$ is the $d$-dimensional pooled input representation produced by the encoder with parameters $\theta$, and $\mathbf{w}\in\mathbb{R}^{d}$ and $b\in\mathbb{R}$ are the learnable classification-head parameters. The sigmoid function $\sigma$ maps the logit to $p_t$, the estimated probability that the target turn merits retention given its context. The admission threshold $\tau_g$ determines whether the turn proceeds to extraction: the indicator function $\mathbf{1}[\cdot]$ returns 1 when the condition holds and 0 otherwise.

The Gate is trained on binary memory-admission labels obtained from consensus among multiple LLM teachers and released annotations. It learns admission directly rather than predicting a semantic memory type, using class-weighted binary cross-entropy. At inference time, preceding messages provide context for the admission decision, while only the admitted target user turn is passed to the extractor. Supervision construction and the training objective are detailed in Appendices~\ref{app:annotation}--\ref{app:gate_input}; training settings and context-window selection are reported in Appendices~\ref{app:config} and~\ref{app:window}.

\subsection{Semantic Dependency-Guided Memory Extraction}
\label{sec:sdg_extract}
We use HanLP~\citep{he2021hanlp}, an NLP toolkit, to obtain tokenization, part-of-speech tags, named entities, and semantic dependencies for each admitted user turn. Based on these analyses, deterministic rules divide the turn into clauses and select predicates with informative arguments and modifiers. For a candidate predicate $p$, the score is
\begin{equation}
 S(p)=\sum_{r\in\mathcal R}\alpha_r I_r(p)+B(p)-P(p),
 \label{eq:pred_score}
\end{equation}
where $I_r(p)$ indicates the presence of an outgoing relation in semantic role group $r$, $\alpha_r$ is its fixed weight, and $B$ and $P$ encode content-cue bonuses and structural penalties. Role groups cover patients or objects, actors, time, purpose/reason/condition, and location. Appendix~\ref{app:extraction} details selection and realization; numerical settings are collected in Appendix~\ref{app:config}.

For each selected predicate, the extractor follows semantic dependencies to identify its arguments and modifiers, extending the text span when needed to retain important entities and quantities. It then takes the corresponding contiguous span from the source, preserving the original word order. Deterministic rules convert first-person expressions into user-attributed statements and restore missing subjects or omitted source content where possible. These operations require no generative LLM calls.
Related or overlapping units may be merged to preserve context, and units exceeding the per-turn limit are combined rather than discarded. A coverage check measures how much content from the cleaned source remains in the extracted text. If coverage is insufficient or no usable unit is produced, the extractor attempts to retain a normalized version of the cleaned source instead. This fallback reduces information loss, although some turns may still yield no entry. The extracted units are concatenated in order into $m_t=\operatorname{Concat}(f_{t,1},\ldots,f_{t,n_t})$, where \(f_{t,i}\) denotes the \(i\)-th unit extracted from turn \(t\) and \(n_t\) is the number of units. The resulting text is embedded and stored as a single memory entry with its timestamp and source-turn provenance. Only the target user turn supplies memory content; preceding dialogue informs the Gate’s admission decision but is not included in the extraction input.

\subsection{Offline Memory Consolidation}
\label{sec:offline_update}
After online memory construction, SLICE-Mem performs offline consolidation to identify redundant or outdated information among stored entries~\citep{diekelmann2010memory,rasch2013sleep}. It updates entry statuses to guide subsequent retrieval while preserving the original text, embeddings, and timestamps, without deleting or rewriting memories.

The procedure begins with deterministic rules for clear duplicates and changes in mutually exclusive personal attributes. Duplicate detection considers whether entries refer to the same occurrence, so similar descriptions of separate events remain available. State changes are processed in timestamp order, allowing a return to an earlier value to be retained as a new transition.
For entries that remain active, the system retrieves similar candidates whose message timestamps are no later than the target’s. A LLM then classifies the relation between the target and each candidate. The four possible decisions are \texttt{ignore}, which leaves the entries unchanged, \texttt{dedup}, which marks the later duplicate as redundant, \texttt{supersede}, which marks the older entry as outdated and links it to its replacement, and \texttt{qualify}, which records a compatible relation while retaining both entries.
Before applying a decision that would exclude an entry from ordinary retrieval, the system checks the stated relation, its scope, and supporting text from the target. Unsupported decisions are ignored. Additional checks enforce consistent status changes and replacement links. The prompt also directs the model to retain complementary information and avoid replacing an entire entry when only part of its content conflicts.
Subsequent retrieval normally uses active entries. Questions matching predefined temporal or state-change cues can additionally retrieve outdated entries and follow their replacement links, with historical information explicitly marked. Because excluding an entry may also hide useful evidence, we evaluate the online and consolidated stores separately. Appendix~\ref{app:consolidation} provides the detailed rules, candidate-selection procedure, validation checks, and historical retrieval conditions.

\section{Experiments}
\label{sec:experiments}
\subsection{Experimental Setup}
\label{sec:experimental_setup}

\paragraph{Benchmarks and baselines.}
We evaluate on Memora Weekly~\citep{uddin2026memora} and LongMemEval-S~\citep{wu2025longmemeval}, comparing SLICE-Mem with LangMem~\citep{langchain2025langmem}, A-MEM~\citep{xu2025amem}, Mem0~\citep{chhikara2025mem0}, MemOS~\citep{li2025memos}, and LightMem~\citep{fang2026lightmem}. The quarterly subset of Memora contributes Gate supervision, while the weekly subset is used for end-to-end evaluation. Supervision construction and split handling are described in Appendix~\ref{app:annotation}; model configurations and evaluation settings are detailed in Appendix~\ref{app:config}.

\paragraph{Experimental settings.}
We evaluate memory construction and consolidation using 8B and 30B Qwen backbones, while fixing answer generation to Qwen3-30B-A3B-Instruct-2507 and judging to GPT-OSS-20B. The construction backbones are used only by LLM-dependent components and their paired consolidators. SLICE-Mem Online and the variant without the Salience Gate do not use a generative construction backbone. For each benchmark, we reuse one fixed evaluation per applicable method across the 8B and 30B blocks. Consolidation and the generative ablations are evaluated separately with their respective construction backbones. The Gate uses 2000 hard-labeled examples, with 1600 for training and 400 for validation, and is independently evaluated on 150 human-annotated turns. Its context window is selected using validation loss. We evaluate a nine-point grid of admission thresholds, $\tau_g\in\{0.1,0.2,\ldots,0.9\}$, using the complete memory-and-QA pipeline on Memora Weekly. The setting $\tau_g=0.2$ attains the highest overall FAMA in this sweep and is used for the main comparisons; Appendix~\ref{app:threshold} reports all nine settings. The same Gate configuration is applied to LongMemEval-S without dataset-specific tuning. All five baseline frameworks are evaluated on the benchmark inputs. Detailed configurations, data-overlap handling, and component evaluations are provided in Appendices~\ref{app:experimental}--\ref{sec:appendix_gate_eval}.

\paragraph{Evaluation metrics.}
On Memora Weekly, we report Forgetting-Aware Memory Accuracy (FAMA), which evaluates remembering, reasoning, and recommending while penalizing reliance on information that should be forgotten. On LongMemEval-S, we report answer accuracy (Acc.) across questions involving temporal reasoning, multi-session evidence, knowledge updates, and information from individual speakers. We measure operational cost using LLM token consumption, construction-stage LLM calls, and reported memory-construction runtime. Token counts cover input and output tokens for summarization, updates and answer generation (QA), and call counts cover memory construction and consolidation. For consolidated systems, both online construction and offline consolidation are included in the reported costs. Metric definitions and detailed accounting procedures are provided in Appendix~\ref{app:accounting}.

\subsection{Main Results}
\label{sec:main_results}

Tables~\ref{tab:memora_main} and~\ref{tab:lmes_e2e} compare effectiveness and efficiency on Memora Weekly and LongMemEval-S under the 30B and 8B construction settings, with answer generation and judging held fixed. We report the online and consolidated states separately. All costs reported for the consolidated state include both online memory construction and offline consolidation; total LLM tokens additionally include answer generation. The online scores are from separate QA runs rather than matched reruns of the consolidated states; the comparison protocol and its limits are described in Appendix~\ref{app:accounting}.

\paragraph{Memora Weekly.}

At the $\tau_g=0.2$ operating point selected by the nine-threshold end-to-end sweep (Appendix~\ref{app:threshold}), SLICE-Mem's separate main-table evaluation under the shared 30720-token QA context achieves 46.60 FAMA in both the 30B and 8B comparisons. This is 0.08 points below MemOS in the 30B block and 3.22 points above MemOS in the 8B block. These results are obtained without generative LLM calls for online admission or extraction. In the 30B comparison, SLICE-Mem uses 0.536M total LLM tokens and reports 0.362 h of parsing-plus-embedding time, compared with 7.009M tokens and 2.378 h for MemOS, reducing token consumption by 92.4\% and the compared time values by 84.8\%. Relative to LightMem, the online pipeline reduces total tokens by 79.2\% and 83.4\% and reported construction time by 64.5\% and 76.0\% in the 30B and 8B comparisons, respectively. These are comparisons of recorded implementation-stage timings, not uniformly instrumented end-to-end latency; Appendix~\ref{app:accounting} details the accounting boundary.

After consolidation, FAMA is 46.74 with the 30B consolidator and 47.29 with the 8B consolidator. The corresponding combined construction-and-consolidation times are 0.874 and 0.819 h, with 2532 consolidation calls in each setting. Total token consumption is 3.356M and 3.340M, respectively. Both consolidated states have lower reported runtime than LightMem, but both use more total tokens than LightMem's 2.584M and 3.231M in the paired settings. Relative to the shared online score of 46.60, FAMA increases by 0.14 after 30B consolidation and by 0.69 after 8B consolidation. Thus, the strongest efficiency gains arise from the online pipeline, while consolidation adds cost and has scale-dependent effectiveness.

\begin{table}[!t]

\caption{Effectiveness and stage-wise cost on Memora Weekly. Summary, Update, and QA report input/output tokens in millions (M); QA denotes answer generation. Bold and underlined values indicate the best and second-best effectiveness or cost within each block. SLICE-Mem Online and w/o Salience Gate share one fixed evaluation per method across blocks. – indicates no value for the metric and + Consolidation includes the online and offline stages.}
\label{tab:memora_main}
\begin{center}
\footnotesize
\setlength{\tabcolsep}{2pt}
\resizebox{\textwidth}{!}{%
\begin{tabular}{@{}lcccccccccc@{}}
\toprule
\multicolumn{1}{c}{\multirow[c]{2}{*}{Method}} & \multirow[c]{2}{*}{FAMA $\uparrow$} & \multicolumn{2}{c}{Summary (M)} & \multicolumn{2}{c}{Update (M)} & \multicolumn{2}{c}{QA (M)} & \multirow[c]{2}{*}{Total (M) $\downarrow$} & \multirow[c]{2}{*}{Calls $\downarrow$} & \multirow[c]{2}{*}{Time (h) $\downarrow$} \\
\cmidrule(lr){3-4} \cmidrule(lr){5-6} \cmidrule(lr){7-8}
 & & In & Out & In & Out & In & Out & & & \\
\midrule
\multicolumn{11}{c}{Qwen3-30B-A3B-Instruct-2507} \\
        \midrule
        LangMem & 24.05 & 5.339 & 0.675 & 6.484 & 1.067 & 1.929 & \underline{0.057} & 15.551 & 3039 & 10.639 \\
        A-Mem & 35.67 & 0.858 & 0.193 & 4.985 & 0.984 & 2.999 & \textbf{0.048} & 10.067 & 2732 & 3.777 \\
        Mem0 & 43.08 & 13.780 & 0.562 & \textbf{0.234} & \textbf{0.010} & \textbf{0.371} & 0.068 & 15.025 & 1551 & 1.856 \\
        MemOS & 46.68 & 5.279 & 1.028 & – & – & 0.633 & 0.069 & 7.009 & 5499 & 2.378 \\
        LightMem & 36.93 & 1.035 & 0.141 & \underline{0.876} & \underline{0.043} & 0.428 & 0.061 & 2.584 & 2939 & 1.019 \\
        \hdashline
        \textsc{SLICE-Mem} & 46.60 & \textbf{0} & \textbf{0} & – & – & 0.474 & 0.062 & \underline{0.536} & \textbf{0} & \textbf{0.362} \\
        + Consolidation & \underline{46.74} & \textbf{0} & \textbf{0} & 2.301 & 0.517 & 0.474 & 0.064 & 3.356 & 2532 & 0.874 \\
        w/o Salience Gate & 43.61 & \textbf{0} & \textbf{0} & – & – & 0.457 & 0.059 & \textbf{0.516} & \textbf{0} & 0.761 \\
        w/o Extractor & \textbf{48.02} & \underline{0.631} & \underline{0.115} & – & – & 0.409 & 0.062 & 1.217 & \underline{1351} & \underline{0.442} \\
        w/o Both & 44.64 & 1.156 & 0.143 & – & – & \underline{0.407} & 0.061 & 1.767 & 1555 & 0.542 \\
        \midrule
        \multicolumn{11}{c}{Qwen3-8B} \\
        \midrule
        LangMem & 29.84 & 5.190 & 0.744 & 5.543 & 0.580 & 1.224 & \underline{0.052} & 13.333 & 3108 & 8.830 \\
        A-Mem & 35.03 & 0.981 & 0.143 & 5.086 & 0.483 & 2.999 & \textbf{0.048} & 9.740 & 3092 & 3.265 \\
        Mem0 & 41.49 & 13.680 & 0.585 & \textbf{0.169} & \textbf{0.008} & \textbf{0.275} & 0.063 & 14.780 & 1551 & 2.583 \\
        MemOS & 43.38 & 4.859 & 0.909 & – & – & 0.532 & 0.067 & 6.368 & 5841 & 4.594 \\
        LightMem & 37.60 & 1.056 & 0.255 & \underline{1.358} & \underline{0.062} & 0.441 & 0.059 & 3.231 & 3686 & 1.509 \\
        \hdashline
        \textsc{SLICE-Mem} & \underline{46.60} & \textbf{0} & \textbf{0} & – & – & 0.474 & 0.062 & \underline{0.536} & \textbf{0} & \underline{0.362} \\
        + Consolidation & \textbf{47.29} & \textbf{0} & \textbf{0} & 2.311 & 0.492 & 0.474 & 0.064 & 3.340 & 2532 & 0.819 \\
        w/o Salience Gate & 43.61 & \textbf{0} & \textbf{0} & – & – & 0.457 & 0.059 & \textbf{0.516} & \textbf{0} & 0.761 \\
        w/o Extractor & 44.05 & \underline{0.629} & \underline{0.100} & – & – & 0.431 & 0.061 & 1.221 & \underline{1345} & \textbf{0.358} \\
        w/o Both & 46.48 & 1.171 & 0.239 & – & – & \underline{0.430} & 0.059 & 1.899 & 1566 & 0.835 \\
        \bottomrule
    \end{tabular}%
}
\par\smallskip
\end{center}
\end{table}

\paragraph{LongMemEval-S.}
The online pipeline achieves 56.8\% accuracy in the 30B comparison, while it matches A-MEM and is 2.5 percentage points below MemOS, the highest-scoring baseline in Table~\ref{tab:lmes_e2e} at 59.3\%. In the 8B comparison, it exceeds the strongest listed 8B baseline, A-MEM at 56.2\%, by 0.6 percentage points. The online result is shared across these comparisons because SLICE-Mem's online construction does not depend on a generative backbone. Its 0.913M total tokens consist entirely of answer-generation input and output, and its reported construction time is 31.679 h. Relative to MemOS in the 30B comparison, these represent reductions of 99.5\% in total tokens and 61.0\% in construction time. Relative to A-MEM in the 8B comparison, the reductions are 99.6\% and 27.3\%, respectively. The Gate, dependency parser, and embedding model still incur computation; eliminating generative construction calls therefore does not imply zero construction cost.

The consolidated state reaches 56.0\% accuracy with both backbones, compared with 56.8\% for the online evaluation. The 30B consolidated state uses 57.650M total tokens and 59762 construction/consolidation calls, 44.7\% and 51.9\% below LightMem, respectively, while achieving 5.2 percentage points higher accuracy. The 8B consolidated state uses 57.728M total tokens and requires 37.984 h for online construction and offline consolidation combined. Compared with A-MEM in the 8B setting, it uses 74.0\% fewer tokens and 12.8\% less reported construction time, with accuracy 0.2 percentage points lower. Substantial cost savings therefore remain possible after consolidation, although neither consolidated LongMemEval-S run exceeds the online score. Additional category-wise results are provided in Appendix~\ref{sec:appendix_taskwise}.

\begin{table}[!t]

\caption{Effectiveness and stage-wise cost on LongMemEval-S. Summary, Update, and QA report input/output tokens in millions (M); QA denotes answer generation. Bold and underlined values indicate the best and second-best effectiveness or cost within each block.}
\label{tab:lmes_e2e}
\begin{center}
\footnotesize
\setlength{\tabcolsep}{0.8pt}
\resizebox{\textwidth}{!}{%
\begin{tabular}{@{}lcccccccccc@{}}
\toprule
\multicolumn{1}{c}{\multirow[c]{2}{*}{Method}} & \multirow[c]{2}{*}{Acc. (\%) $\uparrow$} & \multicolumn{2}{c}{Summary (M)} & \multicolumn{2}{c}{Update (M)} & \multicolumn{2}{c}{QA (M)} & \multirow[c]{2}{*}{Total (M) $\downarrow$} & \multirow[c]{2}{*}{Calls $\downarrow$} & \multirow[c]{2}{*}{Time (h) $\downarrow$} \\
\cmidrule(lr){3-4} \cmidrule(lr){5-6} \cmidrule(lr){7-8}
 & & In & Out & In & Out & In & Out & & & \\
\midrule
\multicolumn{11}{c}{Qwen3-30B-A3B-Instruct-2507} \\
        \midrule
        LangMem & 23.6 & 127.598 & 11.619 & 142.472 & 24.825 & 3.252 & 0.060 & 309.827 & 47634 & 300.459 \\
        A-Mem & \underline{56.8} & 58.789 & 4.226 & 136.010 & 6.753 & 12.385 & 0.050 & 218.213 & 35311 & 32.883 \\
        Mem0 & 13.8 & 262.676 & \underline{4.078} & \textbf{4.431} & \textbf{0.058} & \textbf{0.280} & \textbf{0.032} & 271.557 & \underline{23848} & \textbf{13.683} \\
        MemOS & \textbf{59.3} & 141.753 & 32.948 & – & – & 0.859 & 0.055 & 175.615 & 89341 & 81.252 \\
        LightMem & 50.8 & \underline{24.270} & 12.028 & 63.460 & \underline{4.003} & \underline{0.486} & 0.048 & 104.295 & 124255 & 50.053 \\
        \hdashline
        \textsc{SLICE-Mem} & \underline{56.8} & \textbf{0} & \textbf{0} & – & – & 0.865 & 0.048 & \textbf{0.913} & \textbf{0} & \underline{31.679} \\
        + Consolidation & 56.0 & \textbf{0} & \textbf{0} & \underline{50.348} & 6.389 & 0.865 & \underline{0.047} & \underline{57.650} & 59762 & 38.878 \\
        \midrule
        \multicolumn{11}{c}{Qwen3-8B} \\
        \midrule
        LangMem & 16.8 & 121.464 & 9.273 & 139.104 & 15.605 & 2.459 & \underline{0.040} & 287.945 & 47265 & 334.503 \\
        A-Mem & \underline{56.2} & 58.884 & \underline{2.610} & 143.716 & \underline{4.320} & 12.385 & 0.051 & 221.966 & 36061 & 43.548 \\
        Mem0 & 10.8 & 260.507 & 3.277 & \textbf{4.137} & \textbf{0.050} & \textbf{0.222} & \textbf{0.030} & 268.222 & \underline{23857} & \textbf{21.244} \\
        MemOS & 55.0 & 98.856 & 18.622 & – & – & 0.727 & 0.051 & 118.257 & 60117 & 112.907 \\
        LightMem & 53.4 & \underline{24.058} & 8.682 & 258.222 & 15.230 & \underline{0.565} & 0.046 & 306.804 & 414413 & 103.821 \\
        \hdashline
        \textsc{SLICE-Mem} & \textbf{56.8} & \textbf{0} & \textbf{0} & – & – & 0.865 & 0.048 & \textbf{0.913} & \textbf{0} & \underline{31.679} \\
        + Consolidation & 56.0 & \textbf{0} & \textbf{0} & \underline{50.587} & 6.228 & 0.866 & 0.048 & \underline{57.728} & 59762 & 37.984 \\
        \bottomrule
    \end{tabular}%
}
\end{center}
\end{table}

Across both benchmarks and the 8B and 30B backbone comparisons, SLICE-Mem's online pipeline maintains competitive task performance while substantially reducing reported memory-construction time and generative LLM cost relative to the strongest listed baseline in each setting.

\subsection{Admission and Extraction Quality}
\label{sec:component_evidence}

\paragraph{Does the Gate retain useful evidence?}
At the deployed operating point, the Gate retains 28.1\% of Memora Weekly user turns, filtering approximately 72\% of candidates before parsing and embedding. The shared online evaluation achieves 46.60 FAMA in both comparison blocks (Table~\ref{tab:memora_main}). Retention alone does not establish evidence quality: the end-to-end threshold scan in Table~\ref{tab:gate_threshold} directly tests how more aggressive filtering affects downstream performance. Together, these results show that substantial filtering can preserve competitive answer quality, while the sensitivity analysis identifies the cost of discarding additional turns.

\paragraph{Independent validation of salience judgments.}
On a disjoint, human-annotated set of 150 turns, the selected Gate configuration achieves an AUPRC of $0.898\pm0.021$ over three training seeds. Human and teacher admission decisions agree on 140 examples (93.3\%). The context window is selected using Gate validation loss, not the human-audit scores. This provides independent evidence for the learned admission signal, although the audit uses one annotator and stratified sampling rather than the natural stream distribution. Context-dependent turns remain harder: a representative model obtains AUPRC 0.844 on those turns versus 0.935 on self-contained turns. Training, sampling, and uncertainty estimates appear in Appendix~\ref{sec:appendix_gate_eval}.

\paragraph{Sensitivity and extraction reliability.}
The earlier threshold-sweep runs reveal a coverage--cost trade-off. Their QA evaluations are separate from the shared online evaluation used in the main tables. Raising $\tau_g$ from 0.2 to 0.9 reduces retained turns from 28.1\% to 13.3\% and reported parsing-plus-embedding time from 0.362 to 0.173 h, but lowers overall FAMA from 48.54 to 42.10 and Reasoning FAMA from 46.67 to 32.67. Within this Memora Weekly sweep, moderate filtering yields higher FAMA than more aggressive rejection. The sweep characterizes the trade-off on the data used for threshold selection. At the deployed $\tau_g=0.2$ operating point (3327 admitted turns), the extractor produces empty outputs on 0.51\% of admitted turns and none of the positive-labeled turns. The heuristic fragment rate is 0.96\%, whole-turn fallback occurs on 6.64\%, and realized strings retain 82.80\% of content tokens from positive turns. These automatic diagnostics characterize the admitted stream at the deployment threshold and support the feasibility of deterministic realization, but they do not establish proposition-level factuality, atomicity, or downstream retrieval utility (Appendix~\ref{sec:appendix_extractor_eval}). The following ablations provide the more direct evidence for the modules' downstream contribution.

\subsection{Ablation and Consolidation Analysis}
\label{sec:ablation}

\paragraph{Contributions of Gate and Extractor.}
Here, \emph{w/o Gate} processes all candidate user turns with the deterministic extractor; \emph{w/o Extractor} retains the Gate but replaces deterministic extraction with LLM-based extraction; \emph{w/o Both} omits the Gate and uses LLM-based extraction. Thus, removing the Extractor does not mean storing raw turns.
In the 30B Memora block, removing both components yields 44.64 FAMA and retaining only the Gate yields 48.02. The shared non-generative evaluations obtain 43.61 with the dependency extractor alone and 46.60 with both components. In the 8B comparison, neither component alone improves on the 46.48 no-component configuration, whereas their combination reaches 46.60. Thus, the effectiveness gains from individual components are not uniform across the paired comparisons. The cost effects are clearer: the Gate-only variants still invoke generative extraction, while the extractor-only variants remove these calls but process all candidate turns. Combining both preserves the zero-call online path and reduces reported runtime from 0.761 to 0.362 h relative to the extractor-only variant. Full training, selection, and diagnostic results are in Appendix~\ref{sec:appendix_gate_eval}.

\paragraph{Joint effectiveness and efficiency.}
Against the configuration without either component, the combined online pipeline has 1.96 points higher Memora FAMA in the 30B comparison while reducing reported construction time from 0.542 to 0.362 h. In the 8B comparison, FAMA changes by only 0.12 points, but runtime falls from 0.835 to 0.362 h and generative construction calls fall from 1566 to zero. The benefit is therefore not solely an increase in answer score: a similar score can accompany a substantial change in the computational path. Conversely, the 8B Gate-only ablation is slightly faster than the combined pipeline (0.358 versus 0.362 h), but has lower FAMA and still requires generative extraction. These comparisons distinguish the cost of filtering turns from the cost of representing admitted content.

\paragraph{Task dependence of consolidation.}
The 8B Memora results illustrate why an aggregate improvement should be interpreted cautiously. Relative to the online evaluation, the consolidated state raises Recommending FAMA by 0.77 points and Reasoning by 2.00, while Remembering decreases by 0.71; overall FAMA rises by 0.69. With the 30B consolidator, overall FAMA increases by 0.14, with improved Remembering, unchanged Reasoning, and lower Recommending. The choice of online versus consolidated memory should therefore consider both the target task and the added cost. These are descriptive comparisons across runs, without an estimate of repeated-run variation.

\paragraph{Consolidation and evidence coverage.}
In the 30B Memora comparison, Recommending FAMA changes from 50.23 to 46.34 after consolidation, while Reasoning remains at 43.33. On LongMemEval-S, Multi-Session accuracy changes from 66.92\% to 66.17\% with either consolidator. Knowledge-Update remains at 74.36\% with 30B but changes from 74.36\% to 73.08\% with 8B; Single-Assistant changes from 25.00\% to 23.21\% with 30B and stays at 25.00\% with 8B. These descriptive differences do not isolate which relation decisions caused an answer to change. The low Single-Assistant score also highlights the user-turn-only admission boundary. Appendix~\ref{sec:appendix_taskwise} gives the detailed results.

\section{Discussion and Limitations}
\label{sec:discussion}
SLICE-Mem separates the cost of forming useful memories online from the cost and risks of later reconciliation. Against the five baseline frameworks in Tables~\ref{tab:memora_main} and~\ref{tab:lmes_e2e}, the results show that selective admission and deterministic extraction can preserve competitive downstream performance with substantially fewer generative calls and tokens. They do not establish a universal runtime advantage, and the total cost of a consolidated system must include its offline stage. Component training and teacher annotation are additional preparation costs outside the reported operational comparisons.

Consolidation is not uniformly beneficial. Status-based suppression can restrict retrieval of useful historical evidence even when the original text remains stored. The Memora 30B and 8B results differ, motivating policies that preserve complementary episodes and defer uncertain relation decisions. The evaluation measures downstream effectiveness and operational cost, but does not isolate the effects of individual relation decisions or quantify changes in store compactness.

The present pipeline admits only user turns, limiting coverage of assistant-authored evidence. The one-annotator Gate audit and automatic extractor diagnostics at the deployed $\tau_g=0.2$ do not establish proposition-level factuality or complete coreference resolution. The fixed HanLP parser and extraction rules have not been systematically stress-tested under domain shift, disfluency, or controlled parser errors. The threshold is selected from a full nine-point end-to-end sweep on Memora Weekly, also used for reporting. Exact-string de-duplication does not rule out semantic overlap. Evaluation is limited to two dialogue-memory benchmarks, without heterogeneous-source testing such as RHELM \citep{zhang2026rhelm}.

\section{Conclusion}
We presented SLICE-Mem, a lightweight memory framework that separates online memory formation from offline consolidation. Learned salience-based admission and deterministic dependency-guided extraction eliminate generative LLM calls from online admission and extraction while preserving useful conversational evidence. Experiments on Memora Weekly and LongMemEval-S with 8B and 30B Qwen backbones for memory construction and consolidation demonstrate a favorable effectiveness--cost trade-off. Against MemOS in the 30B comparison, the online pipeline achieves 46.60 versus 46.68 FAMA on Memora and 56.8\% versus 59.3\% accuracy on LongMemEval-S, with 92.4\% and 99.5\% fewer total LLM tokens, respectively. Reported Memora construction runtime decreases by 84.8\%. The consolidated state has task- and backbone-dependent effects on Memora and scores below the online state on LongMemEval-S. These findings support selective, non-generative memory formation as an efficient foundation for persistent-memory agents. Future work will focus on evidence-preserving consolidation and extending coverage to assistant-authored information.

\subsection*{AI use statement}
Generative AI tools were used to edit the manuscript for grammar and readability and to assist with code generation. The authors manually reviewed and verified the AI-assisted code for correctness. The authors reviewed all AI-assisted wording and take full responsibility for the final implementation and manuscript, including its claims and results.

\subsection*{Reproducibility statement}
The Method and Experiments sections describe the SLICE-Mem pipeline, benchmarks, baselines, evaluation metrics, and main comparisons. Appendices A--D provide the annotation and extraction rules, model and hardware settings, evaluation and cost-accounting protocols, parameter-selection analyses, and detailed results. The complete code project has been uploaded as supplementary material to enable readers to reproduce the reported experiments.

\clearpage
\bibliography{slice_references}
\bibliographystyle{plainnat}

\clearpage
\appendix
\section{Methodology Details}
\label{app:method}

\subsection{Salience Annotation and Supervision Construction}
\label{app:annotation}
\paragraph{Memory admission.}
We distinguish a turn’s semantic type from its value for persistent memory. The four auxiliary types are personal semantic information (identity, preferences, habits, and enduring constraints), prospective information (plans, intentions, deadlines, and commitments), autobiographical episodic information (bounded personal events), and non-memory information. The binary admission label indicates whether retaining the target turn’s contribution may improve later continuity, personalization, factual consistency, or plan completion beyond the immediate exchange. A turn may mention personal information without contributing useful persistent evidence, whereas a short context-dependent reply may revise a preference or plan. Admission therefore depends on the target turn’s potential future utility rather than its semantic category alone. Under the annotation protocol, however, the \texttt{non\_memory} type requires a negative admission label. Auxiliary semantic types support corpus balancing and annotation checks; the student learns only binary admission.

\paragraph{Consensus supervision.}
Each of three teachers receives the same rubric and the target turn together with up to $K_a$ preceding messages. The request separates \texttt{[CONTEXT]} from \texttt{[TARGET]} and includes anonymous speaker identifiers. Context is used only to resolve references, ellipsis, alternatives, contrasts, and relative time in the target; facts appearing solely in context are not attributed to the target. The annotation window $K_a$ and the student window $k$ serve different purposes and need not be equal. Each response contains a primary memory type, context-dependence label, binary admission decision, confidence, one to three verbatim target excerpts, and a short rationale. Local validation checks field types, admissible labels, confidence in $[0,1]$, consistency of \texttt{non\_memory} with rejection, and target-grounded evidence. Case-only evidence mismatches are normalized before validation. Invalid responses can receive a correction request containing validation feedback and the target text. An example enters the consensus pool only if all three teachers return valid outputs and agree on binary admission; agreement on auxiliary fields is not required. Released Memora labels are used directly, without teacher relabeling. The actual teacher rubric is Chinese; Appendix~\ref{app:teacher_prompt} presents the system prefix and an English translation of the rubric.

\paragraph{Gate supervision data.}
We construct the Gate supervision corpus from two sources of hard labels, totaling 2000 examples with 1600 for training and 400 for validation. The first source contains 1500 examples from five conversational datasets with complementary domains: Multi-Session Chat (MSC)~\citep{xu2022msc}, ESConv~\citep{liu2021esconv}, Conversation Chronicles~\citep{jang2023chronicles}, Life Event Dialog (LED)~\citep{chen2023led}, and Taskmaster-2~\citep{byrne2020taskmaster2}. Each dataset contributes 300 examples, split into 240 training and 60 validation examples. Candidate user turns are independently annotated by DeepSeek-V4-Pro, GLM-5.2, and Kimi-K2.6 using the protocol above. The second source contains 500 hard-labeled examples derived directly from the released \texttt{share\_memory} annotations in the quarterly subset of Memora~\citep{uddin2026memora}. We sample 50 examples for each of ten personas, with 40 for training and 10 for validation, and keep each conversation group within a single split. User turns of at most 60 characters that exactly match a user utterance in the weekly or monthly subsets are discarded and replaced from the remaining pool to reduce the over-representation of shared boilerplate. The overlap filter targets short exact matches across Memora subsets; it does not constitute a complete audit of semantic overlap. The quarterly subset is used for supervision and the weekly subset for downstream evaluation.

\subsection{Gate Input and Training Objective}
\label{app:gate_input}
The student uses the marked context--target format in Equation~\ref{eq:gate_admission}, with up to $k$ immediately preceding messages and no future messages. Speakers are mapped to anonymous identifiers. Let $y_i\in\{0,1\}$ denote a training label from either unanimous teacher supervision or released Memora annotations, and let $p_i$ denote its predicted admission probability. The objective and positive-class weight are
\begin{equation}
\mathcal L_{\mathrm{gate}}=-\frac{1}{N}\sum_{i=1}^{N}\left[w_+y_i\log p_i+(1-y_i)\log(1-p_i)\right],
\qquad w_+=\operatorname{clip}\!\left(\frac{N_-}{N_+},1,w_{\max}\right),
\label{eq:gate_loss}
\end{equation}
where $N$ is the training-set size and $N_+$ and $N_-$ are its positive and negative counts. The bounded weight compensates for class imbalance. The thresholded deployment rule is given in Equation~\ref{eq:gate_admission}; training configuration and validation-based window selection appear in Appendices~\ref{app:config} and~\ref{app:window}.

\subsection{Dependency-Based Extraction Rules}
\label{app:extraction}
The extractor applies deterministic selection and realization to HanLP tokenization, part-of-speech tags, named entities, and semantic dependencies.

\paragraph{Predicate selection.}
Each role group contributes at most once to Equation~\ref{eq:pred_score}. Bonuses favor content cues, identifiable actors, and first-person actors. Penalties apply to nominally embedded predicates, objectless light verbs, predicates without informative outgoing relations, and selected matrix predicates with nested actions. Candidates are processed by decreasing score, then decreasing span length, then increasing predicate-token index. Hard filters additionally remove restrictive or nominal-modifier candidates, objectless light verbs, candidates lacking informative arguments, and negative-scoring candidates. Integrity checks reject very short realizations, incomplete copular fragments without content cues, discourse-only fragments, and realizations omitting a dollar amount present in the clause. Up to $K_c$ propositions are selected per clause. Overlapping candidates can expand an existing span. A first-person attitude predicate from \emph{want, like, love, prefer}, or \emph{hate} can replace an already selected nested action, preserving the attitude--action relation. For example, \emph{I want to learn Japanese} becomes \emph{User wants to learn Japanese}, retaining the intention rather than asserting that the user already learns the language. Appendix~\ref{app:config} gives all selection weights and limits.

\paragraph{Span construction and realization.}
Permitted semantic relations guide argument, nested-clause, and noun-phrase expansion. Spatial, manner, accompaniment, and related modifiers provide additional attachments. Named entities, proper nouns, numbers, monetary patterns, and selected disconnected content tokens can widen the span. Realization uses the contiguous interval from the minimum to maximum retained token index, including intervening tokens. Token order is preserved within each span; separate units selected from the same clause may follow score order. First-person expressions are normalized to user-attributed statements. Subject detection distinguishes complete subjects from possessives, incomplete verb phrases, and object-control constructions; a default subject is supplied when needed. Requests can be rendered as \emph{User asks to\ldots}, and otherwise unreconstructable fragments as \emph{User noted:\ldots}. Recovery can append missing entities, quoted material, or monetary expressions. These operations use fixed surface rules and make no generative LLM calls.

\paragraph{Merging, coverage, and storage.}
Clauses are processed in source order. If multiple units coexist with an anaphoric or contrast cue, they are merged into one. Exceeding the turn limit $K_t$ also merges all accumulated units rather than retaining only the highest-scoring $K_t$. Otherwise, up to $K_t$ units remain. The adapter joins returned units in extraction order and stores one embedded entry per source turn. Coverage is measured against the source surface after discourse-prefix removal. A token occurrence counts as retained when its lowercase form appears as a substring of the joined lowercase realization. Repeated occurrences count separately; matching uses neither lemmatization nor word boundaries. If the retained fraction is below $\rho$ and the cleaned surface has at least $L_{\min}$ whitespace-separated words, a normalized realization of that surface replaces the extraction. The same surface fallback is attempted when no usable clause unit is produced. Empty or recognized discourse-only inputs can still yield no memory, and the production adapter returns no units after an extraction exception. The safeguard therefore reduces content loss without guaranteeing an entry for every admitted turn. No generative fallback is used.

\subsection{Offline Consolidation Procedure}
\label{app:consolidation}
Consolidation retains entry text and vectors while changing status and relation metadata. The active, outdated, and redundant states correspond to the implementation labels \texttt{active}, \texttt{superseded}, and \texttt{dropped}; the last label does not imply physical deletion.

\paragraph{Deterministic alignment.}
Exact-text comparison strips an initial date prefix, applies Unicode NFKC normalization and lowercasing, collapses whitespace, and trims boundary punctuation. Equal normalized bodies are grouped by shared source-unit provenance when available, or by message timestamp otherwise; event-like entries lacking source-unit provenance are not deduplicated by this rule. This avoids collapsing separate purchases, step records, or meetings merely because they use the same wording. The rule-based state pass handles single-clause residence, relationship status, and digit-valued children counts in timestamp order. A repeated current value is deduplicated, whereas a changed value supersedes its predecessor only when positive timestamps establish a strict order. Processing the sequence preserves a return to an earlier value as a new state transition. Preferences are not treated as mutually exclusive by this pass.

\paragraph{Candidate construction and decisions.}
After rule processing, each active target with an embedding retrieves up to $K_u$ nearest-neighbor hits with timestamps no later than its own. Self matches, inactive entries, candidates below the cosine-similarity threshold, and unordered pairs already assigned to a job are removed. At most $N_u$ remaining neighbors are retained by decreasing similarity. Thus $K_u$ limits initial retrieval and $N_u$ limits the filtered candidate list.

The classifier returns \texttt{ignore}, \texttt{dedup}, \texttt{supersede}, or \texttt{qualify}. Deduplication retains the earlier copy and marks the later one redundant; supersession marks the older entry outdated and records reciprocal replacement links. Qualification relates compatible entries without hiding either. A proposed \texttt{dedup} or \texttt{supersede} must also specify its basis, a concrete entity-and-event or attribute scope, and a short verbatim anchor from the target. The validator checks that the anchor occurs in the target and that the basis supports the requested action. An episodic duplicate additionally requires a same-event basis plus either shared source provenance or an explicit shared event date; supersession requires an explicit event correction or an identifiable current-state change with a strict message-time order. Unsupported hiding decisions are ignored. The prompt in Appendix~\ref{app:consolidation_prompt} distinguishes independent episodes from changes to the same state or event and instructs the model to ignore uncertain or partial conflicts.

Candidate jobs are collected before classification in the store's returned entry order; they are not explicitly sorted chronologically or rebuilt after status changes. Action-specific checks skip redundant entries and inapplicable supersessions. Unparseable JSON yields no relation decisions for that request, and the failure is counted; this occurred twice in the 8B LongMemEval-S update. Unknown or forbidden rewriting actions map to \texttt{ignore}, IDs outside the candidate list are skipped, and aliases such as \texttt{duplicate} are normalized to \texttt{dedup}. Model-service failures remain errors rather than silently invoking rule-only consolidation. Status writes are checked for cycles, unintended reactivation, and inconsistent replacement links, then validated against the final stored state. The output contains relation metadata, never rewritten memory text.

\paragraph{Historical retrieval.}
A case-insensitive lexical pattern recognizes temporal or state-change queries, including expressions such as \emph{when}, \emph{previously}, \emph{used to}, \emph{how long}, \emph{currently}, and \emph{changed}. Purely aggregative queries do not trigger history expansion unless they also match this pattern. For requested retrieval budget $R$, the routine fetches up to $\max(3R,R+16)$ similarity-ranked hits, always excludes redundant entries, and excludes outdated entries unless the trigger matches. It retains at most $R$ eligible hits in similarity order; historical-query packing does not explicitly prioritize active entries. Triggered queries additionally follow direct supersession links from the initially packed entries, with at most $\max(R,8)$ linked-entry lookup attempts. Expansion is one hop rather than recursive, and included IDs are not duplicated. Historical text is marked \texttt{[OUTDATED; superseded]}. The routine returns at most $R+\max(R,8)$ entries for triggered queries and at most $R$ otherwise.

\lstdefinestyle{sliceprompt}{
  basicstyle=\normalfont\normalsize,
  lineskip=0pt,
  columns=fullflexible, keepspaces=true,
  breaklines=true, breakatwhitespace=false,
  showstringspaces=false, tabsize=2,
  aboveskip=8pt, belowskip=8pt,
  frame=single, framerule=0.4pt, framesep=6pt,
  xleftmargin=6.4pt, xrightmargin=6.4pt,
  literate={→}{{$\rightarrow$}}1 {—}{{---}}1 {–}{{--}}1
           {€}{{\texteuro}}1 {£}{{\pounds}}1 {’}{{'}}1 {“}{{``}}1 {”}{{''}}1
}
\newtcolorbox{sliceprose}{breakable,colback=white,colframe=black!45,boxrule=0.35pt,arc=0pt,left=5pt,right=5pt,top=4pt,bottom=4pt,before skip=3pt,after skip=5pt,fontupper=\normalfont\normalsize,before upper={\justifying\setlength{\parindent}{0pt}\setlength{\parskip}{2pt}\emergencystretch=2em}}
\subsection{Teacher Annotation Prompt}
\label{app:teacher_prompt}

\begin{sliceprose}
\textbf{Task:}You are an independent annotation teacher. Follow the rubric below and return JSON only.\par
\textbf{Conversational Memory Annotation Guideline:}\par
Decide what core proposition the TARGET contributes for a general conversational memory system. CONTEXT may be used only to resolve deixis, ellipsis, contrasts, options, or relative time in TARGET. Information that appears only in CONTEXT is not part of TARGET’s contribution. Do not rely on dataset labels, test-set idiosyncrasies, class quotas, sentence length, or keywords, and do not make unsupported inferences.\par
Make three judgments separately: 1. \textnormal{primary\_memory\_type}: what type of proposition TARGET contributes; 2. \textnormal{context\_dependency}: whether the same proposition is recoverable from TARGET alone; 3. \textnormal{candidate\_memory}: whether storing the proposition may help later turns or sessions.\par
Matching a memory type does NOT automatically imply \textnormal{candidate\_memory=true}. The student gate is trained ONLY on \textnormal{candidate\_memory}.\par
\textbf{Decision procedure:}\par
\textbf{1. Recover TARGET’s proposition}\par
Identify what TARGET asserts, confirms, corrects, selects, commits to, or updates. Judge TARGET’s contribution, not the whole dialogue topic. If TARGET’s main act is asking, thanking, greeting, acknowledging, or reacting emotionally, treat attached clauses as secondary by default unless they clearly assert a storable fact, preference, plan, or commitment.\par
\textbf{2. Choose exactly one `primary\_memory\_type`}\par
Only these four types are allowed:\par
- \textnormal{personal\_semantic}: stable or generalizable knowledge about the TARGET speaker (identity, relationships, preferences, habits, abilities, values, long-term constraints/states). Usually not tied to a single occasion. - \textnormal{autobiographical\_episodic}: a bounded personal event/experience at a specific occasion, situation, time, or place. Describes what happened once; the event need not be worth storing. - \textnormal{prospective}: future intentions, plans, goals, commitments, appointments, deadlines, or expected actions attributable to the TARGET speaker. Corrections of future plans remain \textnormal{prospective}. Shared future arrangements the speaker commits to also count as \textnormal{prospective} when the main update is a future commitment. - \textnormal{non\_memory}: TARGET contributes no storable personal proposition (greetings, thanks, acknowledgments, generic questions, small talk, pure restatement, unsupported guesses, rumor-like chatty evaluations aimed at questioning/chitchat, or only confirming/repeating already-settled CONTEXT content without adding durable personal facts or future commitments).\par
Operational split: stable person knowledge \ensuremath{\rightarrow} \textnormal{personal\_semantic}; one bounded event \ensuremath{\rightarrow} \textnormal{autobiographical\_episodic}; speaker’s future action/commitment \ensuremath{\rightarrow} \textnormal{prospective}; else \ensuremath{\rightarrow} \textnormal{non\_memory}.\par
If multiple types seem to apply, pick by primary future value with priority: 1) speaker’s future plan/commitment (\textnormal{prospective}); 2) else stable person knowledge (\textnormal{personal\_semantic}); 3) else bounded event (\textnormal{autobiographical\_episodic}). Do not upgrade a one-off experience to \textnormal{personal\_semantic} unless TARGET itself states a generalizable preference/habit/value/long-term state. Do not treat casual interest, rumors, or “I heard X is good” chitchat as stable preferences.\par
\textbf{3. “Remove CONTEXT” test}\par
- \textnormal{self\_contained}: TARGET alone yields the same core proposition and the people/events/options/time needed to store it correctly. - \textnormal{context\_dependent}: without CONTEXT, the core proposition is incomplete, ambiguous, or clearly changes. If correct storage needs CONTEXT to resolve pronouns, ellipsis, which plan/order/option is meant, or who “they/you/it” refers to, label \textnormal{context\_dependent}. Pronouns/ellipsis/short replies are strong cues but not automatic. If CONTEXT is \textnormal{(none)}, do not invent missing meaning.\par
\textbf{4. Decide `candidate\_memory`}\par
Set \textnormal{true} only if storing TARGET’s contribution may improve later continuity, personalization, factual consistency, relevant follow-ups, or completion of unfinished plans/commitments, AND that benefit goes beyond the next few minutes of the current exchange.\par
Set \textnormal{false} for: - small talk, thanks, greetings, acknowledgments, and generic questions; - question-dominated turns even if an interest topic is mentioned; - rumor/shallow evaluations that only sustain chitchat; - actions that only serve the next few minutes and then expire (e.g., pickup/delivery for this order, “I’m almost there”, momentary commands); - trivial one-off events and experiences with no lasting value; - vague events whose key content is underspecified; - facts that appear only in CONTEXT; - interpretations that require unsupported inference.\par
Travel/work/daily anecdotes default to \textnormal{false} unless TARGET already frames them as durable person knowledge or an unfinished commitment.\par
\textbf{5. Calibration examples}\par
- \textnormal{(none)} / \textnormal{I usually prefer quiet hotels.} \ensuremath{\rightarrow} \textnormal{personal\_semantic}, \textnormal{self\_contained}, \textnormal{true} - \textnormal{Are you still moving on Friday?} / \textnormal{No, next Monday.} \ensuremath{\rightarrow} \textnormal{prospective}, \textnormal{context\_dependent}, \textnormal{true} - \textnormal{(none)} / \textnormal{I had toast this morning.} \ensuremath{\rightarrow} \textnormal{autobiographical\_episodic}, \textnormal{self\_contained}, \textnormal{false} - \textnormal{For future trip plans, shall we use metric units?} / \textnormal{Yes, let's keep metric units.} \ensuremath{\rightarrow} \textnormal{prospective}, \textnormal{context\_dependent}, \textnormal{true} - \textnormal{Have you been to Paris?} / \textnormal{Not yet; I went to Brittany and liked the countryside. Going with friends?} \ensuremath{\rightarrow} \textnormal{autobiographical\_episodic}, \textnormal{self\_contained}, \textnormal{false} - \textnormal{Any new books?} / \textnormal{I heard Kristin Hannah is very good; have you heard of her?} \ensuremath{\rightarrow} \textnormal{non\_memory}, \textnormal{self\_contained}, \textnormal{false} - \textnormal{Your order will be ready in about 20 minutes.} / \textnormal{Great. I'll be there to get it.} \ensuremath{\rightarrow} \textnormal{prospective}, \textnormal{context\_dependent}, \textnormal{false} - \textnormal{Do your parents pressure you?} / \textnormal{I just don't want to disappoint them. They work so hard for me.} \ensuremath{\rightarrow} \textnormal{personal\_semantic}, \textnormal{context\_dependent}, \textnormal{true}\par
\textbf{6. Output}\par
Return exactly one JSON object (no Markdown, no extra fields):\par
\{ "primary\_memory\_type": "prospective", "context\_dependency": "context\_dependent", "candidate\_memory": true, "confidence": 0.91, "evidence": ["verbatim substring copied from TARGET"], "reason": "One short TARGET-based reason, using CONTEXT only if needed." \} Memory type must be one of the four names above. \textnormal{context\_dependency} must be \textnormal{self\_contained} or \textnormal{context\_dependent}. \textnormal{candidate\_memory} must be a JSON boolean. \textnormal{confidence} is a number in [0, 1]. \textnormal{evidence} has 1–3 non-empty substrings copied character-for-character from TARGET (including case and punctuation). \textnormal{reason} is one short evidence-based sentence. Reason internally; output JSON only.\par
\end{sliceprose}

\subsection{Consolidation Relation Prompt}
\label{app:consolidation_prompt}
\nopagebreak
\begin{sliceprose}
You classify relations between memories of ONE user's conversations. Never rewrite, merge, or delete memory text. Output JSON only.\par
TARGET has a message timestamp; CANDIDATES are earlier or equal-time messages. Message time is NOT necessarily event time. Similar topic, different amount, or later timestamp alone does not imply replacement.\par
Actions:\par
ignore: unrelated, uncertain, different entities/scopes/events, or only part of a multi-fact memory conflicts.\par
qualify: compatible information; keep both.\par
dedup: clearly the SAME fact/event retold, not another occurrence. Two purchases, daily step records, meetings, or attendance lists normally coexist even with identical wording. A return to a previous state is a new change, not a duplicate.\par
supersede: TARGET explicitly corrects the SAME event, or changes the SAME current-state attribute of the SAME entity and scope. Budgets, goals, project fields, plans and schedules may change. Different days' spending/steps are independent events. Daily spending is not a budget. Preserve historical evidence. If hiding a whole entry loses unrelated facts, ignore.\par
For ignore/qualify output only id and action. For dedup/supersede also output:\par
basis: same\_fact|same\_event|event\_correction|state\_change\par
scope: a short specific entity+attribute/event identifier (include project/category/period when relevant)\par
anchor: a SHORT verbatim quote from TARGET supporting the identity or explicit change; do not fabricate a quote.\par
Use dedup only with same\_fact/same\_event; supersede only with event\_correction/state\_change. When identity or scope is ambiguous, ignore. Do not infer an event correction from a new purchase or a different day.\par
Schema: \{ "decisions": [ \{ "id": "candidate id", "action": "ignore" \}, \{ "id": "candidate id", "action": "supersede", "basis": "state\_change", "scope": "user weekly lunch budget", "anchor": "change my weekly lunch budget" \} ] \}\par
Keep output compact; no explanations outside these fields.
\end{sliceprose}

\section{Experimental Details}
\label{app:experimental}

\subsection{Models, Hardware, and Hyperparameters}
\label{app:config}
\paragraph{Gate training.}
We instantiate the Gate encoder with ModernBERT-base~\citep{warner2025modernbert}. Speakers are mapped to anonymous identifiers, and the $k$ messages immediately preceding the target turn are taken from the current dialogue. Sequences are left-truncated to 512 tokens so that the target and its nearest context are retained. We set $w_{\max}$=10, giving a positive-class weight of approximately 1.40, use AdamW with a learning rate of $2\times10^{-5}$, weight decay 0.01, warmup ratio 0.1, batch size 4, and four-step gradient accumulation. Models are trained for at most 10 epochs, evaluated every 50 steps, and checkpointed according to validation binary cross-entropy.

\paragraph{Teacher requests.}
The annotation context contains at most $K_a=5$ preceding messages. Requests use temperature 0, JSON-object output, disabled thinking, and a maximum generated length of 400 tokens. The request loop permits two retries after the initial attempt. The student context window is selected separately as described in Appendix~\ref{app:window}.

\paragraph{Extractor configuration.}
For Equation~\ref{eq:pred_score}, the weights for patient/object, actor, temporal, purpose/reason/condition, and locative roles are 3, 2, 2, 2, and 1, respectively. A configured content cue contributes a single bonus of 3.5; an identifiable actor adds 0.5 and a first-person actor adds another 0.5. Penalties are 1.5 for a nested matrix predicate, 6 for restrictive or nominal-modifier embedding, 2 for an objectless light verb, and 4 for missing informative arguments. The associated hard filters are detailed in Appendix~\ref{app:extraction}. The per-clause and per-turn limits are $K_c=2$ and $K_t=2$; exceeding $K_t$ merges the accumulated units. Nested-clause and noun-phrase traversal depths are capped at 5 and 4. Near-duplicate suppression uses an overlap threshold of 0.85 relative to the smaller token set. Coverage fallback uses $\rho=0.55$ and $L_{\min}=8$ whitespace-separated words in the discourse-stripped source. Coverage counts token occurrences through case-insensitive substring matching, as defined in Appendix~\ref{app:extraction}. Additional splitting at comma--conjunction boundaries applies to segments with at least 16 words.

\paragraph{Consolidation configuration.}
Initial neighbor retrieval uses $K_u=20$, followed by a minimum cosine similarity of 0.45 and a retained-candidate limit of $N_u=10$. A request therefore contains one target and at most ten eligible neighbors. Relation-classifier requests use temperature 0, top-$p$ 1, seed 42, JSON-object output, and disabled thinking. The configurable generation cap defaults to 1024 output tokens and bounds the relation-decision JSON for each request; it is not a length limit on the memory store or on rewritten memory text.

\paragraph{End-to-end models and hardware.}
Published baseline numbers are not reused because their hardware, model endpoints, and accounting conventions differ from ours. All methods are rerun on the same benchmark inputs, using official implementations and method-specific configurations where available. We use Qwen3-30B-A3B-Instruct-2507 as the answer-generation model for every system and GPT-OSS-20B as the fixed judge. For LLM-dependent memory construction, we evaluate Qwen3-30B-A3B-Instruct-2507 and Qwen3-8B; offline consolidation uses the model scale paired with the corresponding construction setting. The Salience Gate and dependency extractor are unchanged across these settings. We standardize the evaluation prompt and, where architectures permit, the retrieval budget, while preserving method-specific construction logic. API calls, token usage, and construction-stage runtimes are collected from the evaluation-run records. Experiments are conducted on a server with two NVIDIA RTX 4090 GPUs and two NVIDIA A100 GPUs. The Gate window experiments use the three seeds specified in Appendix~\ref{app:window}; all other randomized data processing uses seed 42.

\subsection{Evaluation Protocol and Cost Accounting}
\label{app:accounting}
\paragraph{Metrics.}
On Memora, Forgetting-Aware Memory Accuracy (FAMA) combines memory retention and forgetting at the question level. For question $q$, let $N_{p,q}$ and $N_{f,q}$ be the numbers of memory-presence and forgetting-absence criteria, and let $C_{p,q}$ and $C_{f,q}$ count satisfied criteria. Define $\mathrm{MPA}_q=C_{p,q}/N_{p,q}$ and $\mathrm{FAA}_q=C_{f,q}/N_{f,q}$. The evaluator sets MPA to zero when there are no presence criteria and FAA to one when there are no forgetting criteria. It computes
\begin{equation}
\mathrm{FAMA}_q=\max\!\left(0,\mathrm{MPA}_q-\lambda_q(1-\mathrm{FAA}_q)\right),\qquad
\lambda_q=\frac{N_{f,q}}{N_{p,q}+N_{f,q}},
\end{equation}
with $\lambda_q=0$ when both counts are zero. We average question-level FAMA within each persona, then average the ten persona scores and multiply by 100. Task-specific FAMA uses the same procedure restricted to questions of that task. The diagnostic MPA and FAA instead pool the corresponding criteria within each persona before averaging persona scores. On LongMemEval-S, answer accuracy is the proportion of correctly answered questions. Open-ended answers are graded by GPT-OSS-20B using a fixed evaluation prompt.

We report three operational cost measures. \emph{Calls} counts generative LLM invocations during memory construction and consolidation, excluding answer generation and judging. Token consumption is separated into Summary, Update, and QA input/output tokens. Main-table token counts are in millions (M), summed across the evaluated personas or groups; Total includes all three stages and excludes judging. Totals are calculated from unrounded counts. LightMem's construction costs include its online and offline update stages. \emph{Runtime} reports construction-stage time recorded in the evaluation runs, with offline reconciliation added for consolidated rows. These measurements describe the recorded memory-processing stages rather than the latency of answer generation and judging. In particular, Memora SLICE-Mem Online reports parsing-plus-embedding time, whereas baseline timings follow their recorded construction workflows; these are not uniformly instrumented end-to-end wall-clock measurements. Using the same server does not eliminate implementation, batching, or scheduling effects. For LongMemEval-S, we report 31.679 h for the online pipeline but do not report a parser/embedding/I/O breakdown, so the dominant bottleneck cannot be inferred from the aggregate. Time reductions are calculated from the original unrounded timings and displayed hours are rounded to three decimals. Teacher annotation and Gate training are preparation costs outside these operational totals. QA token consumption is reported separately from construction calls and runtime; zero construction calls therefore does not imply zero answer-generation cost. Retention ratios characterize admission selectivity, while downstream effectiveness is measured by FAMA or answer accuracy. Main-table comparisons do not include independent repeated-run variance or significance tests. The three-seed Gate experiments are not repeated end-to-end evaluations; small answer-score differences should therefore be read descriptively.

\section{Component Evaluation and Parameter Selection}
\label{sec:appendix_gate_eval}
The same Gate, admission threshold, and extraction rules are used for both downstream benchmarks. We distinguish training-set supervision, validation-based model selection, independent human auditing, and operating-point characterization on Memora Weekly.

\subsection{Training and Context-Window Selection}
\label{app:window}
\paragraph{Context-window selection and Gate evaluation.}
Using at most five preceding messages and identical data and optimization settings, we compare $k\in\{1,3,5\}$ using seeds $\{20260726,20260727,20260728\}$. Classification reports use a fixed threshold of 0.5. At inference we score only user turns and admit a turn when $p_t\geq\tau_g$. The window is selected using mean validation loss; operating-threshold selection is reported separately in Appendix~\ref{app:threshold}. 
Table~\ref{tab:gate_validation} reports results on the 400-example Gate validation set used for early stopping and checkpoint selection. The mean validation loss selects $k=1$. The human results in Appendix~\ref{app:human} are reported as an independent evaluation rather than as a selection criterion.

\begin{table}[htbp]

    \caption{Salience Gate performance on the validation set ($n=400$). Results are mean $\pm$ standard deviation over three seeds at threshold 0.5.}
    \label{tab:gate_validation}
\begin{center}
    \small
    \resizebox{0.84\textwidth}{!}{%
    \begin{tabular}{ccccc}
        \toprule
        $k$ & Loss & F1 & Precision & Recall \\
        \midrule
        1 & 0.382$\pm$0.015 & 0.853$\pm$0.013 & 0.841$\pm$0.015 & 0.866$\pm$0.027 \\
        3 & 0.399$\pm$0.033 & 0.845$\pm$0.022 & 0.823$\pm$0.051 & 0.870$\pm$0.011 \\
        5 & 0.428$\pm$0.035 & 0.828$\pm$0.017 & 0.863$\pm$0.052 & 0.800$\pm$0.063 \\
        \bottomrule
    \end{tabular}%
    }
\end{center}
\end{table}

\subsection{Independent Human Evaluation}
\label{app:human}
We construct an independent human-annotated set of 150 user turns from the five conversational datasets, with 30 examples per source. Sampling is stratified by context dependence and teacher agreement. The set is disjoint from the Gate training and validation data. One annotator independently applies the same admission criteria, yielding 84 positive examples. This set is used only for component-level evaluation and is excluded from training, early stopping, context-window selection, and operating-threshold selection. We report AUPRC and fixed-threshold precision, recall, F1, and macro-F1. Confidence intervals for window comparisons use paired bootstrap resampling of the same 150 examples (10000 resamples), with predictions averaged across seeds.

\paragraph{Independent human evaluation.}
Table~\ref{tab:gate_human_gold} reports performance on the independent 150-example human-annotated set. Values are means and standard deviations over three training seeds, and the reporting threshold is fixed at 0.5. The human annotations are not used for training, early stopping, context-window selection, or operating-threshold selection. Using paired bootstrap resampling on the same 150 examples and predictions averaged across seeds, $k=1$ improves AUPRC over $k=3$ by 0.043, with a 95\% confidence interval of $[0.007,0.085]$. Relative to $k=5$, the AUPRC difference is 0.026 and its confidence interval includes zero, whereas the F1 difference is 0.100 and its confidence interval excludes zero. Human and teacher admission decisions agree on 140 of 150 examples (93.3\%). For a representative $k=1$ seed, AUPRC is 0.935 on context-self-contained turns and 0.844 on context-dependent turns. The human set has 84 positive examples (56\%) because it is stratified for auditing rather than sampled to match the natural class distribution. Its fixed-threshold scores therefore characterize this audit distribution and are not estimates of accuracy on the full conversational stream.

\begin{table}[htbp]

    \caption{Salience Gate performance on the independent human-annotated set ($n=150$). Results are mean $\pm$ standard deviation over three seeds at the fixed reporting threshold 0.5.}
    \label{tab:gate_human_gold}
\begin{center}
    \small
    \resizebox{\textwidth}{!}{%
    \begin{tabular}{cccccc}
        \toprule
        $k$ & AUPRC  & Precision  & Recall  & F1  & Macro-F1 \\
        \midrule
        1 & 0.898$\pm$0.021 & 0.820$\pm$0.043 & 0.746$\pm$0.050 & 0.780$\pm$0.026 & 0.763$\pm$0.028 \\
        3 & 0.851$\pm$0.029 & 0.826$\pm$0.030 & 0.750$\pm$0.074 & 0.785$\pm$0.044 & 0.770$\pm$0.037 \\
        5 & 0.876$\pm$0.006 & 0.833$\pm$0.041 & 0.595$\pm$0.062 & 0.692$\pm$0.030 & 0.703$\pm$0.012 \\
        \bottomrule
    \end{tabular}%
    }
\end{center}
\end{table}

\subsection{Admission-Threshold Selection and Sensitivity}
\label{app:threshold}

False rejections remove evidence from all subsequent processing, so the sweep includes high-recall operating points. We evaluate the complete memory-and-QA pipeline at all nine thresholds $\tau_g\in\{0.1,0.2,\ldots,0.9\}$ on the same Memora Weekly stream of 11860 candidate user turns. We use $\tau_g=0.2$ in the main comparisons because it yields the highest overall FAMA in this end-to-end sweep. LongMemEval-S uses the same threshold without separate tuning. Reporting all nine settings makes the selection transparent but does not provide an independent test of the selected operating point on Memora Weekly; a separate QA run on the same benchmark does not create a held-out dataset. Table~\ref{tab:gate_threshold} reports downstream performance and computational cost for all nine settings. At $\tau_g=0.2$, the sweep obtains 48.54 FAMA, retains 28.1\% of candidate turns, and records 0.362 h of parsing-plus-embedding time. For Tables~\ref{tab:memora_main} and~\ref{tab:memora_taskwise}, we re-evaluate this selected configuration and obtain 46.60 FAMA for the online state. The $\tau_g=0.8$ point reuses a prior full run rather than an independent rerun within the sweep. Higher thresholds reduce runtime further but lower overall FAMA; at $\tau_g=0.9$, Reasoning FAMA drops from 46.67 to 32.67. Thus, the selected threshold prioritizes answer quality while substantially reducing the stream processed by the extractor, rather than minimizing runtime alone.

\begin{table}[htbp]

    \caption{End-to-end sensitivity of memory curation performance and computational cost to the Salience Gate threshold $\tau_g$. Rem., Rea., and Rec. denote task-specific FAMA for remembering, reasoning, and recommending, respectively. Keep is the percentage of candidate user turns retained for memory construction.}
    \label{tab:gate_threshold}
\begin{center}
    \small
    \resizebox{\textwidth}{!}{%
    \begin{tabular}{ccccccccc}
        \toprule
        $\tau_g$ & FAMA & Rem. & Rea. & Rec. & MPA  & FAA & Keep (\%) & Runtime (h)  \\
        \midrule
        0.1 & 46.81 & 45.28 & 47.33 & 47.82 & 60.14 & 49.47 & 34.6 & 0.432 \\
        0.2 & 48.54 & 50.27 & 46.67 & 48.67 & 60.70 & 53.73 & 28.1 & 0.362 \\
        0.3 & 48.07 & 46.72 & 48.67 & 48.82 & 61.44 & 53.52 & 25.0 & 0.312 \\
        0.4 & 47.30 & 49.05 & 46.67 & 46.19 & 60.31 & 55.77 & 22.8 & 0.296 \\
        0.5 & 46.99 & 46.73 & 46.00 & 48.23 & 59.34 & 54.40 & 20.9 & 0.263 \\
        0.6 & 46.89 & 49.87 & 43.33 & 47.46 & 60.22 & 56.34 & 19.3 & 0.247 \\
        0.7 & 44.87 & 45.02 & 41.33 & 48.26 & 59.54 & 54.41 & 17.6 & 0.225 \\
        0.8 & 44.13 & 47.94 & 38.00 & 46.44 & 59.68 & 52.47 & 15.8 & 0.195 \\
        0.9 & 42.10 & 45.20 & 32.67 & 48.42 & 56.00 & 57.96 & 13.3 & 0.173 \\
        \bottomrule
    \end{tabular}%
    }
\end{center}\end{table}

\subsection{Extractor Diagnostics}
\label{sec:appendix_extractor_eval}

At the deployment operating point $\tau_g=0.2$, the Gate admits 3327 Memora Weekly user turns, including 1608 with positive memory-admission labels. Using HanLP v4.2.1 with the main-pipeline extraction configuration, empty outputs occur on 0.51\% of admitted turns and none of the positive turns. A heuristic fragment flag occurs on 0.96\% of admitted turns (0.25\% of positive turns), and 6.64\% of turns use whole-turn fallback. Content-token retention on positive turns is 82.80\% over 18986 source content tokens, with 1.08 intermediate units per admitted turn on average. These diagnostics characterize extractor behavior at the deployment threshold, but they are automatic proxy metrics rather than proposition-level human judgments and do not establish factuality, atomicity, cross-turn reference resolution, retrieval utility, or answer correctness.

\section{Additional Results and Analysis}
\label{sec:appendix_taskwise}
The following breakdowns complement the aggregate comparisons in Section~\ref{sec:main_results}. Memora FAMA and LongMemEval-S accuracy use different definitions and should not be compared numerically across benchmarks.

\subsection{Task-wise Results on Memora}
\label{app:memora_tasks}
\begin{table}[htbp]
\caption{Task-wise results on Memora Weekly. Rem., Rea., and Rec. denote Remembering, Reasoning, and Recommending. Scores are reported on a 0--100 scale. SLICE-Mem Online results are shared across blocks.}
\label{tab:memora_taskwise}
\begin{center}
\small
\setlength{\tabcolsep}{3pt}
\resizebox{\textwidth}{!}{%
\begin{tabular}{lccccccccc}
\toprule
\multicolumn{1}{c}{\multirow[c]{2}{*}{Method}} & \multirow[c]{2}{*}{\shortstack{Overall\\FAMA}} & \multicolumn{3}{c}{Rem.} & \multicolumn{2}{c}{Rea.} & \multicolumn{3}{c}{Rec.} \\
\cmidrule(lr){3-5}\cmidrule(lr){6-7}\cmidrule(lr){8-10}
 & & FAMA & MPA & FAA & FAMA & MPA & FAMA & MPA & FAA \\
\midrule
\multicolumn{10}{c}{Qwen3-30B-A3B-Instruct-2507} \\
\midrule
LangMem & 24.05 & 20.63 & 27.89 & 86.88 & 10.67 & 17.78 & 40.85 & 54.61 & 68.72 \\
A-Mem & 35.67 & 62.68 & 74.71 & 76.39 & 10.00 & 16.67 & 34.33 & 49.01 & 61.61 \\
Mem0 & 43.08 & 51.23 & 60.87 & 66.02 & 23.33 & 36.67 & 54.68 & 82.83 & 35.42 \\
MemOS & 46.68 & 58.35 & 64.74 & 60.89 & 25.33 & 40.00 & 56.37 & 76.98 & 42.76 \\
LightMem & 36.93 & 39.82 & 47.50 & 66.67 & 22.67 & 37.78 & 48.30 & 76.10 & 39.16 \\
\hdashline
\textsc{Slice-Mem} ($\tau_g=0.2$) & 46.60 & 46.24 & 56.56 & 67.30 & 43.33 & 50.00 & 50.23 & 77.69 & 35.05 \\
\quad + Consolidation & 46.74 & 50.55 & 58.70 & 67.79 & 43.33 & 52.22 & 46.34 & 78.01 & 29.33 \\
\midrule
\multicolumn{10}{c}{Qwen3-8B} \\
\midrule
LangMem & 29.84 & 34.43 & 40.06 & 72.24 & 10.67 & 17.78 & 44.43 & 53.64 & 65.32 \\
A-Mem & 35.03 & 60.47 & 74.10 & 71.76 & 9.33 & 15.56 & 35.30 & 51.65 & 65.21 \\
Mem0 & 41.49 & 50.83 & 57.77 & 74.22 & 27.33 & 41.11 & 46.32 & 75.66 & 32.83 \\
MemOS & 43.38 & 51.39 & 57.97 & 61.81 & 27.33 & 41.11 & 51.42 & 76.70 & 37.56 \\
LightMem & 37.60 & 42.90 & 51.73 & 65.94 & 21.33 & 35.56 & 48.57 & 76.06 & 34.72 \\
\hdashline
\textsc{Slice-Mem} ($\tau_g=0.2$) & 46.60 & 46.24 & 56.56 & 67.30 & 43.33 & 50.00 & 50.23 & 77.69 & 35.05 \\
\quad + Consolidation & 47.29 & 45.53 & 56.89 & 68.76 & 45.33 & 53.33 & 51.00 & 79.11 & 34.45 \\
\bottomrule
\end{tabular}%
}
\end{center}
\end{table}

Table~\ref{tab:memora_taskwise} reports the shared online score of 46.60 FAMA in both blocks. With the 30B consolidator, overall FAMA reaches 46.74. Remembering changes from 46.24 to 50.55, Reasoning remains at 43.33, and Recommending changes from 50.23 to 46.34. With the 8B consolidator, overall FAMA reaches 47.29. Remembering changes from 46.24 to 45.53, Reasoning from 43.33 to 45.33, and Recommending from 50.23 to 51.00. These cross-run comparisons do not isolate individual consolidation decisions or estimate repeated-run variability.

\subsection{Category-wise Results on LongMemEval-S}
\label{app:lmes_categories}
\begin{table}[htbp]

    \caption{Category-wise accuracy (\%) on LongMemEval-S in the 30B and 8B construction comparisons. Parentheses report benchmark category sizes. Answer generation and judging are fixed across methods. SLICE-Mem Online results are shared across blocks.}
    \label{tab:lmes_categorywise}
\begin{center}
    \small
    \resizebox{\textwidth}{!}{%
    \begin{tabular}{lccccccc}
        \toprule
        \multicolumn{1}{c}{Method} & Acc. & \begin{tabular}[c]{@{}c@{}}Temporal\\($n=133$)\end{tabular} & \begin{tabular}[c]{@{}c@{}}Multi-Session\\($n=133$)\end{tabular}
        & \begin{tabular}[c]{@{}c@{}}Knowledge-\\Update\\($n=78$)\end{tabular} & \begin{tabular}[c]{@{}c@{}}Single-User\\($n=70$)\end{tabular}
        & \begin{tabular}[c]{@{}c@{}}Single-\\Assistant\\($n=56$)\end{tabular} & \begin{tabular}[c]{@{}c@{}}Single-\\Preference\\($n=30$)\end{tabular} \\
        \midrule
        \multicolumn{8}{c}{Qwen3-30B-A3B-Instruct-2507} \\
        \midrule
        LangMem & 23.6 & 12.03 & 15.79 & 47.44 & 37.14 & 23.21 & 16.67 \\
        A-Mem & 56.8 & 36.84 & 51.13 & 69.23 & 77.14 & 98.21 & 13.33 \\
        Mem0 & 13.8 & 6.02 & 8.27 & 16.67 & 15.71 & 44.64 & 3.33 \\
        MemOS & 59.3 & 47.37 & 56.06 & 73.08 & 80.00 & 67.86 & 26.67 \\
        LightMem & 50.8 & 42.86 & 56.39 & 61.54 & 75.71 & 16.07 & 40.00 \\
        \hdashline
        \textsc{SLICE-Mem} & 56.8 & 43.61 & 66.92 & 74.36 & 77.14 & 25.00 & 36.67 \\
        + Consolidation & 56.0 & 42.86 & 66.17 & 74.36 & 77.14 & 23.21 & 33.33 \\
        \midrule
        \multicolumn{8}{c}{Qwen3-8B} \\
        \midrule
        LangMem & 16.8 & 9.77 & 12.78 & 29.49 & 24.29 & 23.21 & 3.33 \\
        A-Mem & 56.2 & 36.84 & 50.38 & 67.95 & 74.29 & 98.21 & 16.67 \\
        Mem0 & 10.8 & 1.50 & 7.52 & 11.54 & 7.14 & 44.64 & 10.00 \\
        MemOS & 55.0 & 44.36 & 45.11 & 70.51 & 88.57 & 66.07 & 6.67 \\
        LightMem & 53.4 & 43.61 & 62.41 & 67.95 & 71.43 & 19.64 & 40.00 \\
        \hdashline
        \textsc{SLICE-Mem} & 56.8 & 43.61 & 66.92 & 74.36 & 77.14 & 25.00 & 36.67 \\
        + Consolidation & 56.0 & 43.61 & 66.17 & 73.08 & 75.71 & 25.00 & 33.33 \\
        \bottomrule
    \end{tabular}%
    }
\end{center}
\end{table}

The online SLICE-Mem state obtains 43.61\% on Temporal, 66.92\% on Multi-Session, and 74.36\% on Knowledge-Update questions. In the 30B comparison, it leads the listed baseline frameworks on Multi-Session and Knowledge-Update, while MemOS has the highest Temporal accuracy (47.37\%). Single-Assistant accuracy is 25.00\%, consistent with the user-turn-only admission boundary. The 30B consolidated state reaches 56.0\% overall. Relative to the online results, Multi-Session changes from 66.92\% to 66.17\%, Temporal from 43.61\% to 42.86\%, and Single-Preference from 36.67\% to 33.33\%; Single-User and Knowledge-Update remain unchanged, while Single-Assistant changes from 25.00\% to 23.21\%. The category counts sum to 500, and their weighted scores reproduce 56.8\% online and 56.0\% after consolidation to rounding precision.

In the 8B comparison, the online state leads the listed baseline frameworks on Multi-Session (66.92\%) and Knowledge-Update (74.36\%). MemOS leads on Temporal and Single-User, A-MEM on Single-Assistant, and LightMem on Single-Preference. The 8B consolidated state reaches 56.0\% overall. Relative to the online result, Temporal and Single-Assistant remain at 43.61\% and 25.00\%; Multi-Session changes to 66.17\%, Knowledge-Update to 73.08\%, Single-User to 75.71\%, and Single-Preference to 33.33\%. The two consolidators have the same overall accuracy here but different category-level outcomes. The reported category sizes and accuracies imply 284 correct answers online and 280 after either consolidation, a net loss of four out of 500. For 30B, the net loss is one answer each in Temporal, Multi-Session, Single-Assistant, and Single-Preference; for 8B, it is one each in Multi-Session, Knowledge-Update, Single-User, and Single-Preference. These are aggregate net changes, not counts of changed predictions or causal attributions to individual consolidation decisions.

\subsection{Extended Ablation and Consolidation Analysis}
\label{app:extended_ablation}
The online ablations in Table~\ref{tab:memora_main} separate admission filtering from non-generative extraction. Relative to the extractor-only variant, combining both reduces reported construction runtime by 52.4\% while retaining zero generative construction calls. The Gate-only variant still requires 1351 and 1345 construction calls in the 30B and 8B comparisons, respectively. Its 8B runtime of 0.358 h is slightly below the combined pipeline's 0.362 h, but the latter achieves higher FAMA without generative extraction calls. For the consolidated runs, measured offline-update time is approximately 0.511 h with 30B and 0.456 h with 8B. Adding the reported online formation time yields the reported 0.874- and 0.819-h combined totals. These costs accompany different task-level effects across the two scales.

\paragraph{Decision reliability and untested failure mechanisms.}
The validators in Appendix~\ref{app:consolidation} check format, references, and textual support; they do not establish the semantic correctness of relation labels. Two unparseable 8B LongMemEval-S decisions are reported there, but this is not an estimate of the semantic error rate. Entry-level suppression may hide complementary content in a multi-fact turn, and lexical historical-query triggering may miss implicit temporal or aggregation needs. These are mechanisms suggested by the design, not audited causes of the observed score changes. Establishing causality would require tracing changed answers to retrieved evidence and the corresponding relation decisions.

\subsection{Positioning and Evaluation Scope}
\label{app:related_scope}
EnSIMem organizes entity--property indexes with source-evidence provenance and adaptive retrieval \citep{meng2026ensimem}; SLICE-Mem instead targets inexpensive turn-level admission and extraction, without an equivalent entity-indexed retrieval design. TSM models event semantic time and durative states \citep{su2026tsm}; message timestamps, status links, and lexical historical retrieval in SLICE-Mem do not implement semantic-time normalization or durative-memory construction. Semantic Anchoring combines dependency structure, discourse information, and coreference with vector memory \citep{chatterjee2025anchoring}; our target-turn extractor uses a narrower deterministic realization mechanism. A-MAC explicitly models admission using interpretable factors and an LLM-assisted utility assessment \citep{zhang2026amac}; explicit admission itself is therefore not unique to SLICE-Mem. Our distinction is a compact learned Gate paired with deterministic extraction, with no generative online construction calls. These are architectural contrasts, not empirical superiority claims over those systems.

RHELM combines dialogues with heterogeneous sources such as documents and emails and tests evolving, cross-source memory \citep{zhang2026rhelm}. Our two-benchmark evaluation does not establish robustness in that setting. Cross-source ingestion, assistant-authored evidence, and implicit state reconstruction remain outside the tested coverage.

\end{document}